\documentclass[12pt]{article}
\usepackage{amsmath}
\usepackage{graphicx,psfrag,epsf}
\usepackage{enumerate}
\usepackage{natbib}
\usepackage{xcolor}

\usepackage[most]{tcolorbox}  
\usepackage{hyperref}
\usepackage{mathtools}
\tcbset{boxmath/.style 2 args={sharp corners, colback=white,
        colframe=#1, remember as=#2, size=fbox}}
\newcommand{\blind}{0}

\begin{document}

\def\spacingset#1{\renewcommand{\baselinestretch}%
{#1}\small\normalsize} \spacingset{1}



{\if0\blind{
\title{\bf Fast parameter estimation of Generalized Extreme Value distribution using Neural Networks}
\author{
    \texorpdfstring{Sweta Rai\textsuperscript{1}, Alexis Hoffman\textsuperscript{2},  Soumendra Lahiri\textsuperscript{3}, Douglas W.~Nychka\textsuperscript{1}\\}{Sweta Rai\textsuperscript{1}, Alexis Hoffman\textsuperscript{2},  Soumendra Lahiri\textsuperscript{3}, Douglas W.~Nychka\textsuperscript{1}}
    \texorpdfstring{Stephan R.~Sain\textsuperscript{2}, and Soutir Bandyopadhyay\textsuperscript{1}}{Stephan R.~Sain\textsuperscript{2}, and Soutir Bandyopadhyay\textsuperscript{1}}
    \vspace{1em}
    \thanks{SR, SB, and DWN’s work has been partially supported by the National Science Foundation, CMMI-2210840, and SL's work has been partially supported by the National Science Foundation, CMMI-2210811. Corresponding author email: sbandyopadhyay@mines.edu.}\\
    \hspace{.2cm}\textsuperscript{1} Department of Applied Mathematics and Statistics, Colorado School of Mines,\\ Golden, CO\\
    \hspace{.2cm}\textsuperscript{2} Jupiter Intelligence, Boulder, CO\\
    \hspace{.2cm}\textsuperscript{3} Washington University in St. Louis, St. Louis, Missouri
}
\date{}\maketitle}\else\fi}

\if1\blind
{
  \bigskip
  \bigskip
  \bigskip
  \begin{center}
    {\LARGE\bf Title}
\end{center}
  \medskip
}\fi

\begin{abstract}
The heavy-tailed behavior of the generalized extreme-value distribution makes it a popular choice for modeling extreme events such as floods, droughts, heatwaves, wildfires, etc. However, estimating the distribution's parameters using conventional maximum likelihood methods can be computationally intensive, even for moderate-sized datasets. To overcome this limitation, we propose a computationally efficient, likelihood-free estimation method utilizing a neural network. Through an extensive simulation study, we demonstrate that the proposed neural network-based method provides Generalized Extreme Value (GEV) distribution parameter estimates with comparable accuracy to the conventional maximum likelihood method but with a significant computational speedup. To account for estimation uncertainty, we utilize parametric bootstrapping, which is inherent in the trained network. Finally, we apply this method to 1000-year annual maximum temperature data from the Community Climate System Model version 3 (CCSM3) across North America for three atmospheric concentrations: 289 ppm $\mathrm{CO}_2$ (pre-industrial), 700 ppm $\mathrm{CO}_2$ (future conditions), and 1400 ppm $\mathrm{CO}_2$, and compare the results with those obtained using the maximum likelihood approach. 
\end{abstract}

{\textit{Keywords:}  Deep neural networks;  Generalized Extreme Value distribution; Parameter estimation; Sufficient statistics; Extreme quantiles.

\newpage\clearpage\pagebreak

\section{Introduction}
\label{sec:intro}
It is widely acknowledged that the Earth is currently undergoing a period of climate change, which is resulting in extreme variability in weather patterns and constant disturbances in the surrounding environment. In order to inform planning and mitigate risks associated with these changes, it is crucial to gain a thorough understanding of these extreme events and quantify their potential impact (see, \cite{adger2003social, power2021decadal}).  
To estimate the probability of extreme events occurring, extreme value theory (EVT) is widely used in various fields such as econometrics (\cite{bali2003generalized}), finance (\cite{broussard1998behavior}), materials science (\cite{castillo2012extreme}), environmental science (\cite{katz2010statistics, martins2000generalized}), and reliability engineering (\cite{smith1991weibull}).
For more detailed information on EVT, please refer to \cite{fisher1928limiting, hosking1985estimation, smith1985maximum, smith1990extreme, bali2003generalized,   coles2001introduction, gumbel2004statistics, davison2015statistics} and the references therein. Under the univariate EVT framework, the GEV distribution is commonly used for modeling extreme events due to its flexibility and sound theoretical foundation. One of its significant characteristics is the shape parameter, which governs various tail behaviors, ranging from a restricted distribution to one with heavy tails, as described in \cite{coles2001introduction}, \cite{gumbel2004statistics}, \cite{davison2015statistics}, \cite{haan2006extreme}.

\ 
Typically, the maximum likelihood (ML) or moment-based methods are utilized to estimate GEV parameters, both of which can be intricate and time-consuming as it involves numerical computation due to domain dependence on the shape, and other parameters. As a result, recent methods for fitting a GEV distribution have looked into alternatives to utilizing the exact likelihood (see, \cite{casson1999spatial, el2009joint, opitz2018inla, erhardt2012approximate, castro2021practical}).
One such approach is the use of Bayesian methods, such as Approximate Bayesian Computation (ABC), which has been employed to examine the dependence for modeling max-stable processes (see, \cite{blum2010approximate, erhardt2012approximate, erhardt2016modelling, sisson2018overview}).
On the other hand, several deep-learning algorithms have recently been developed for likelihood-free inference for extreme events (see, \cite{creel2017neural, lenzi2021neural, sainsbury2022fast, richards2022unifying}, and the references therein). These works show promising results in accuracy as compared to the classical approaches, along with a potential speed-up factor in the overall estimation process. However, the application of these works mostly focuses on modeling spatial extremes using max-stable processes for higher dimensions.

This work is motivated by recent applications of the ABC method and the deep-learning algorithm to spatial extremes as described above, as well as the use of neural networks (NN) for time series and spatial data, as evidenced in papers such as \cite{chon1997linear, cremanns2017deep,  gerber2020fast, wikle2022statistical, majumder2022modeling}. In this work, we present a new estimation method that utilizes a NN to fit univariate GEV distributions to extreme events. It is crucial to note that identifying the marginal GEV distributions is typically required for modeling any multivariate extreme process, and therefore, in this paper, we focus on the computationally efficient modeling of univariate GEV distributions. 
The use of informative statistics instead of full datasets for modeling extremes has been reported in several studies, including \cite{creel2013indirect, creel2017neural, gerber2020fast}. By including both extreme quantiles and the standard quartiles ($\mathrm{Q_1}$, $\mathrm{Q_2}$, and $\mathrm{Q_3}$), our NN is able to effectively capture the important tail behavior in extreme event modeling. The utilization of sample quantiles as inputs for the network is supported by the concept of order statistics. During training, we input the sample quantiles and apply activation functions to generate optimal nonlinear representations, which are then used to estimate the GEV parameters. The model outputs an estimate of the GEV parameters, which define the distribution of the extreme event. To ensure a robust model, we utilized simulated values within a reasonable parameter range for training and selected a large training size. Furthermore, we also utilized a validation set to monitor the model's performance during training.

\vskip1em
\noindent The use of a NN for this problem offers the following benefits.
\begin{enumerate}
\item[(a)] The NN architecture is well suited for inference, allowing for the efficient estimation of parameters. In particular, the network can be quickly evaluated once trained, resulting in significant speed gains of up to 100 - 1000 times compared to traditional methods.
\item[(b)] To address the issue of uncertainty in our parameter estimates, we adopt the bootstrapping approach, which has been widely supported by previous research (see, for example, \cite{cooley2007bayesian, kysely2008cautionary, huang2016estimating, lenzi2021neural, gamet2022flexible, yadav2022flexible, sainsbury2022fast}). 
In particular, to generate confidence intervals, we utilized the parametric bootstrap, which is typically computationally intensive. However, the trained NN enables efficient simulation and evaluation of bootstrap samples, resulting in the rapid generation of confidence intervals.
\end{enumerate}

Finally, our model is employed to analyze the maximum temperature extremes in the CCSM3 model's run at three distinct $\mathrm{CO}_2$ concentrations throughout North America. By examining annual maximum temperature data, we demonstrate the accuracy and advantages of our approach compared to the classical ML method. The fast evaluation speed of the neural estimators facilitated efficient uncertainty quantification through parametric bootstrap sampling. Our findings indicate that we can produce hundreds of spatial confidence intervals within a matter of seconds.

The remainder of the paper is structured as follows. Section \ref{sec:meth} offers an overview of the GEV distribution and elucidates the proposed NN model. Section \ref{sec: sim} showcases the outcomes of our simulation study. Section \ref{sec:casestudy} delves into our CCSM3 runs case study, and lastly, Section \ref{sec:conc} recapitulates our findings, examines the behavior and limitations of our model, and presents our conclusion.

\section{Methods}
\label{sec:meth}
This section provides an overview of the structure of the GEV distribution and outlines our model framework. It also includes information on the approximate statistics chosen as inputs for the network and the network architecture used in our model.

\subsection{Generalized Extreme-Value Distribution}

The GEV distribution, introduced by \cite{jenkinson1955}, with location-scale parameters $(\mu, \sigma)\in$ $\mathcal{R} \times (0, \infty)$ and the shape or tail-index parameter $\xi \in \mathcal{R}$. It has the cumulative distribution function (CDF)  denoted as

\begin{equation*}
    \mathcal{F}(x)=
    \begin{cases}
     \mathrm{exp} \left\{- \left[1+\xi(x-\mu)/\sigma\right]^{-1/\xi} \right\}, & \text{$\mathrm{if}\hspace{0.1 cm} \xi\neq 0$} \vspace{0.1 in}\\   
   
     \mathrm{exp} \left\{-\mathrm{exp}\left[-(x-\mu)/\sigma\right] \right\}, & \text{$\mathrm{if}\hspace{0.1 cm}\xi=0$}
    \end{cases}       
\end{equation*}

The support of $\mathcal{F}$ is determined by the interval 
\begin{equation}
S_{\theta}=\{x\in \mathcal{R}:\sigma +\xi(x-\mu)>0\},
\label{eq:support-constr}
\end{equation}
where $\theta = (\mu, \sigma, \xi)$.
Therefore, the CDF $\mathcal{F}$ is defined only for values of $x$ that fall within $S_{\theta}$. When evaluating the risk associated with extreme events in extreme value analysis (EVA), return levels are a crucial component. These levels estimate the expected values of extreme quantiles that may occur within a specific time frame, or return period, represented by $T$. The equation for computing return levels is given by $z_{T} = \mathcal{F}^{-1}(1-T^{-1})$, where $\mathcal{F}^{-1}$ refers to the inverse CDF of the GEV distribution, and $T$ denotes the return period.
The use of the $(1-T^{-1})$ quantile as the threshold is important because it represents the average frequency with which this threshold is exceeded over the specified return period. 

The GEV distribution can take on three different forms depending on the sign of its shape parameter, $\xi$. These forms are the Gumbel distribution for light-tailed distributions ($\xi=0$), the Fréchet distribution for heavy-tailed distributions ($\xi>0$), and the Weibull distribution for short-tailed distributions ($\xi<0$). The sign of $\xi$ also determines whether the GEV distribution is upper-bounded ($\xi>0$) or lower-bounded ($\xi<0$).

Determining the large-sample asymptotics of the ML estimator (MLE) for the GEV parameters is challenging due to the dependence of the support $S_{\theta}$ on the parameters $\theta$. However, \cite{smith1985maximum} and \cite{bucher2016maximum} have shown that by restricting the lower bound of $\xi > -1/2$, the asymptotic properties can be preserved. Therefore, in our simulation study, we limit the range of $\xi > -1/2$ to ensure the validity of the MLE results.

\subsection{Approximate sufficient statistics}
In this paper, we propose an indirect inference method for estimating the unknown parameter $\theta \in \Theta \subset \mathcal{R}^k$ of the underlying GEV distribution, based on a sample $y= (y_1,y_2,\ldots,y_n )$ and using a minimal set of lower-dimensional statistics as input to a NN. The approach is specifically designed to infer the heavy-tailed behavior of the GEV distribution. It achieves this by utilizing a set of extreme quantiles from both the lower and upper ends, including the $Q_1$, $Q_2$, and $Q_3$.

While the use of summary statistics in indirect inference is a useful approximation method, it may not always achieve optimal asymptotic efficiency. To address this issue, it is important to carefully select informative quantiles that accurately capture the heavy-tailed behavior of the GEV distribution while minimizing the input dimensionality of the NN. To identify the optimal set of extreme quantiles, different choices of quantiles are experimented with, and their impact on the network's behavior is observed. Selecting an informative statistic is crucial for the success of indirect inference as it can impact the computational efficiency, robustness, and interpretability of the NN. Therefore, approximately sufficient statistics are selected to ensure computational efficiency and robustness. 

This approach is motivated by previous studies (\cite{creel2013indirect, jiang2017learning}) that have shown promising results using informative statistics in statistical modeling. Still, to the best of our knowledge, this is the first study that explores the use of extreme quantiles as input to a NN for the estimation of the GEV distribution parameters. The study aims to examine the ability of a given set of quantiles to estimate the parameters of the GEV distribution using a sophisticated deep NN, with the goal of providing a more efficient and accurate method for inferring the heavy-tailed nature of the distribution. Further details about the NN framework are provided in Section \ref{subsec:NNframework}.

\subsection{NN Framework}
\label{subsec:NNframework}
Our estimation technique involves a deep NN that takes the quantile values as its inputs and returns an approximate $\theta=(\mu, \sigma, \xi)$ as the dependent output.
Let  $\mathcal{P}= \mathcal{P}(\textbf{y}) \in \mathcal{R}^m $ be the quantile/percentile values, which is a scalar vector, $m < n$, where m and n are the sizes of $\mathcal{P}(y)$ and $y$ respectively. The function $\mathcal{N}$ maps $\mathcal{P}$ to $\theta$ such that $\widehat\theta = (\widehat\mu, \widehat\sigma, \widehat\xi)$ is obtained as the corresponding estimate

\begin{equation*} 
\mathcal{N} :\mathcal{R}^{m} \to \Theta; \hspace{0.1cm} \mathrm{where} \hspace{0.2cm} \mathcal{N}(\mathcal{P})= \widehat\theta.
\end{equation*}

The feedforward neural network (FFNN) $\mathcal{N}$ has $L$ layers, where the first layer is the input layer, the final layer is the output layer, and the remaining $L-2$ layers are the hidden layers. Let the $j^{th}$ layer of $\mathcal{N}$ has $n_j$ neurons, $j=1,2,\ldots,L$, and is characterized by the non-linear activation function $f_j$. The interaction between the successive layers is defined by the recursive equation
$$h_j = f_j(b_j + w_j h_{j-1}), \quad j=1,2,\ldots,L,$$
where $h_j$ is the vector output for all neurons in the $j^{th}$ layer, $b_j$ is the bias vector, $w_j$ is the weight matrix connecting the $(j-1)^{th}$ layer to the $j^{th}$ layer, and $h_{j-1}$ is the  vector output of the $(j-1)^{th}$ layer. This equation captures the computation performed by each layer of the network, where the input to each layer is the output of the previous layer after being transformed by the weight matrix, bias vector, and activation function. Thus, $\mathcal{N}$ maps inputs to outputs through a sequence of non-linear transformations performed over the subsequent layers. During training, the weights and biases are adjusted to minimize the discrepancy between the predicted $\widehat\theta$ and the true $\theta$ for a given input $\mathcal{P}$. 

Selecting an appropriate performance loss  function to measure the discrepancy between $\theta$ and $\widehat\theta$ is crucial in building a NN algorithm.
A commonly used metric for measuring the performance of a NN algorithm is the mean squared error (MSE) loss, which is computed over a selected batch of the training sample and is expressed as follows:

\begin{equation}
\mathrm{MSE}(\omega)=\dfrac{1}{n_B} \sum_{i=1}^{n_B}\|\theta_i -\widehat\theta_i(\omega)\|_2^2
=\dfrac{1}{n_B} \sum_{i=1}^{n_B} \left\{(\mu_i-\widehat\mu_i(\omega))^2 + (\sigma_i-\widehat\sigma_i(\omega))^2 + (\xi_i-\widehat\xi_i(\omega))^2 \right\},
\label{eq:mse}
\end{equation}

where $n_B$ is the batch size, $\omega$ denotes the matrix that encodes the network's weights and biases for a given iteration stage, and $\|\cdot\|_2 $ is the $l_2$ norm. Minimizing the MSE loss during training is essential for obtaining accurate estimates of $\theta$, thereby achieving a close prediction of the true parameter value. 

However, in the case of estimating the parameters of the GEV distribution, it is important to ensure that the estimated parameters satisfy the support constraint of GEV denoted by \ref{eq:support-constr}.
To achieve the support constraint, we modify the MSE loss function by adding a penalty term that accounts for any violations within the selected batch's sample $y_{n_B}$.
We express the penalty term as $C(y_{n_B}, \widehat\theta_{\mathrm{NN}})$, which returns a value of 1 if any $y_i$ in $y_{n_B}$, $i=1,2,\ldots,{n_B}$, violates the support constraint, and 0 otherwise.

\begin{equation*}
    C(y_{n_B}, \widehat\theta_{NN}) =\begin{cases}
    1, & {\mathrm{if} \hspace{0.1 cm} y_i \hspace{0.1 cm} \mathrm{violates} \hspace{0.1 cm} \mathrm{domain} \hspace{0.1 cm} \mathrm{constraint} \hspace{0.1 cm} \mathrm{for}\hspace{0.1 cm} \mathrm{any} \hspace{0.1 cm} i=1,2,...,n_{B}}\\
    0, & \mathrm{otherwise}.\\
       \end{cases}
\end{equation*}
Therefore, we define a regularized loss function for our problem as follows:
\begin{equation}
\mathrm{Loss}(\omega)= \mathrm{MSE}(\omega) + \lambda \hspace{0.1cm} \hspace{0.1cm} C(y_{n_B}, \widehat\theta_{\mathrm{NN}}(\omega)),
\label{eq:cust-loss}
\end{equation}
where $\lambda$ is a weight that balances the MSE and the penalty term, $C(y_{n_B}, \widehat\theta_{\mathrm{NN}}(\omega))$.

\begin{figure}[h]
\centering
\includegraphics[width=0.8\textwidth]{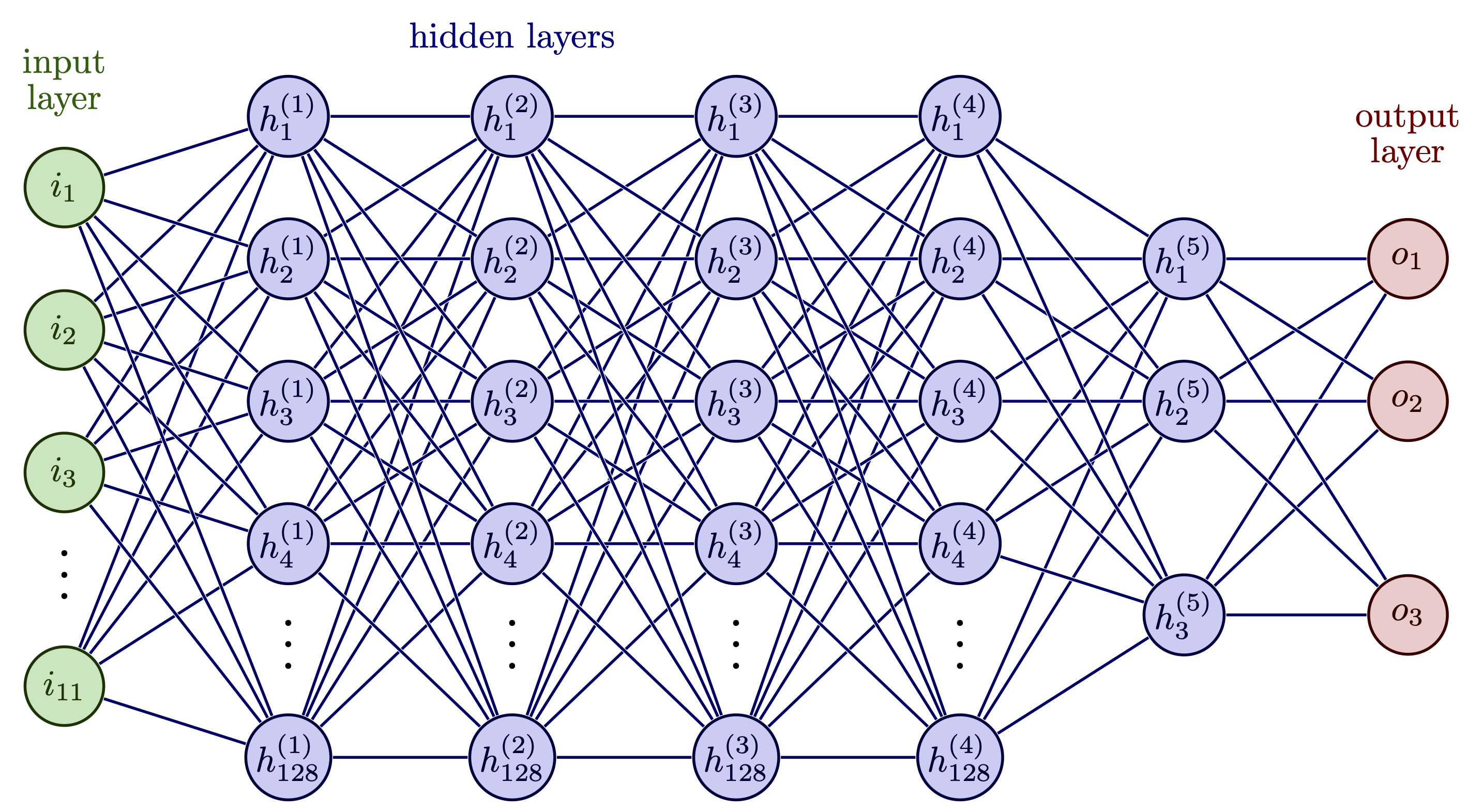}
\caption{Network architecture of the NN model $\mathcal{N}$, taking $\mathcal{P}; \{p_1,p_2,\ldots,p_{11}\}$ as input to the network and returns three scalar values as output.}
\label{fig:1}
\end{figure}

We optimize this loss function using a suitable optimizer ( e.g. we use RMSprop for our study), which involves iteratively updating the weights of the neural network based on the gradients of the loss function. This allows us to train $\mathcal{N}$ to estimate the parameters while ensuring they meet the domain constraint of the GEV.

\subsection{Network Training}

\label{subsec:train}
To train $\mathcal{N}$, we generate a comprehensive training dataset by simulating values from the GEV distribution across a range of feasible parameters $\theta$. Specifically, we generate $340,000$ parameter configurations for training and validation, uniformly sampling  over the range of $\mu \in (1, 50)$, $\sigma \in (0.1, 40)$, and $\xi \in (-0.4, 0.4)$, with the choices of $\mu$ and $\sigma$ informed by our analysis of temperature, precipitation, and wind data. Once a parameter configuration is established, we simulate a GEV sample and further standardize the sample to improve model performance. We achieve standardization by subtracting the sample mean and scaling by the sample interquartile range (IQR), which also rescales the values of $\mu$ and $\sigma$. Notably, the restricted range of $\xi> -1/2$ aligns with the literature, as the ML estimate is valid within this range and supported by the asymptotic property established in prior research \cite{smith1985maximum, bucher2016maximum}.

To serve as the input to the NN, we select a suitable set of percentiles, $\mathcal{P}$. To identify an appropriate set of percentiles, we consider a range of values spanning from the ${0.01}^{th}$ to the ${99.99}^{th}$ percentile over the generated GEV samples given by
\begin{equation}
\mathcal{P}= \left\{{\mathbf{0.01^{th}}, \hspace{0.1cm}\mathbf{0.1^{th}},
\hspace{0.1cm}\mathbf{1^{th}},  \hspace{0.1cm}\mathbf{{10}^{th}},  \hspace{0.1cm} {25}^{th},
\hspace{0.1cm} {50}^{th}, 
\hspace{0.1cm} {75}^{th}, 
\hspace{0.1cm}\mathbf{{90}^{th}},
\hspace{0.1cm}\mathbf{{99}^{th}},
\hspace{0.1cm}\mathbf{{99.9}^{th}}, \hspace{0.1cm}\mathbf{{99.99}^{th}}}\right \},
\end{equation}
where we have boldfaced the extreme percentile values for emphasis. By selecting percentiles from this wide range of values, we can capture the full spectrum of the GEV samples and better understand the heavy-tailed behavior of the distribution. 

\vskip1em
\noindent To train the NN, we explore two scenarios: (1) generating fixed 1000-sized GEV samples across the parameter configuration for training-validation-test, and (2) generating varying-sized GEV samples to better approximate real-world conditions.

\begin{enumerate}
\item[(1)]\label{(1)}
For each parameter configuration, a GEV sample of size $1000$ is simulated. To optimize the simulation process, we employed both vectorization techniques and utilized the GPU support available in Google Colab. This combination allowed us to perform the simulation for $340,000$ configurations, each with a sample size of $1000$, in an efficient manner, taking approximately 12.6 minutes. To facilitate train-validation purposes, we divide the $340,000$ parameter configuration into a training set ($N_{train}$ = 300,000) and a validation set ($N_{valid}$ = 40,000) to monitor overfitting. In addition, we defined a testing set ($N_{test}$ = 10,000) to evaluate the model's behavior over the same parameter ranges mentioned above. Working with a large training set ensures the optimization of the weights involved in the network, resulting in a reliable mapping from inputs to outputs. To prevent overfitting, we employed early stopping measures during the training process, monitoring the validation loss and learning rate. The training process was stopped if there was no improvement in the validation loss.

\item[(2)]
To ensure our study's generalizability to real-world scenarios with limited observations, we investigate how the size of GEV samples used for training affects our findings. For training and validation, we use the same $N_{train}$ and $N_{valid}$ parameter configurations, but to generate GEV samples, we choose sizes ranging from 30 to 1000. During the training and validation phase, a total of $340,000$ parameter configurations are generated, with $68,000$ configurations assigned randomly per size for GEV sample generation. For evaluation, we create a test set by fixing $\mu$ at 0 and generating a $20 \times 20$ regular grid of $(\sigma, \xi) \in (0.1, 40) \times (-0.4, 0.4)$. And then generate GEV samples of sizes ranging across the different sample sizes, each configuration with 100 replications. Our findings demonstrate that increasing the sample size from small to moderate values leads to a significant improvement in estimate accuracy. Refer to Section~\ref{sec: sim} for additional details.
\end{enumerate}

\begin{table}[h!]
\centering
\caption{Summary of the FFNN model}
\begin{tabular}{llll} \hline
\hspace{0.3cm}Layer type &  Output shape &  Activation & Parameters \\ \hline
\hspace{0.5cm}Dense & \hspace{0.3cm}[-, 128] & \hspace{0.4cm}Relu & \hspace{0.4cm}1536\\ 
\hspace{0.5cm}Dense & \hspace{0.3cm}[-, 128] & \hspace{0.3cm} Relu & \hspace{0.3cm} 16512  \\ 
\hspace{0.5cm}Dense & \hspace{0.2cm} [-, 128] & \hspace{0.3cm} Relu & \hspace{0.3cm} 16512  \\ 
\hspace{0.5cm}Dense & \hspace{0.2cm} [-, 128] & \hspace{0.3cm} Relu & \hspace{0.3cm} 16512  \\ 
\hspace{0.5cm}Dense & \hspace{0.2cm} [-, 3] & \hspace{0.3cm} Tanh & \hspace{0.3cm} 387\\ \hline
\end{tabular}\\
 {Note: The input layer has shape [-, 11]; 11 quantile values as inputs, and returning the estimated GEV parameters as output with shape [-, 3].}
\label{table:1}
\end{table}

For both scenarios, we maintain a consistent network architecture and utilize the loss function as described in Eq.~\ref{eq:cust-loss} to train and optimize our model, employing the RMSprop optimizer. The optimizer is initialized with a learning rate of 0.001, allowing efficient and effective adjustments to the model's parameters during training. Table \ref{table:1} provides a comprehensive overview of the architecture, including the output shape per layer and the activation functions employed.  

Our NN's output layer uses a customized activation function that returns three scalar values, corresponding to the shifted $\mu$, shifted $\sigma$, and $\xi$, respectively. We design this activation function by combining the tanh, relu, and tanh activation functions to handle the possible range of these parameters and ensure accurate and reliable estimates. To better model heavy-tailed distributions, we employ the RMSprop compiler for algorithm optimization (see, \cite{hinton2012neural}), updating the weights for every batch of 64 samples. Early-stopping criteria based on validation MSE outlined in Eq.~\ref{eq:mse} is employed to prevent overfitting, stopping the training process when there is no improvement observed in the loss. We save the best weights obtained at the $28^{th}$ epoch. To implement the model, we opt for the fixed sample scenario since it is trained using fixed 1000-GEV samples and is more efficient in terms of accuracy. The model takes 260 seconds to complete one epoch, and training for 150 epochs with early stopping at the $28^{th}$ epoch takes a total of 2.11 hours.

\subsection{Sample Standardization}
\label{subsec:sample-std}
In this section, we will discuss the importance of standardizing the sample before training the network. The standardization process has several benefits, including improved stability, estimation, and performance. However, our main objective here is to make the network more versatile and applicable to a wider range of extreme scenarios by making it invariant to different scales and units of measurement. 

We center and scale the GEV sample using the sample mean and IQR. Let $y=(y_1, y_2, y_3,…, y_n)$ be a sample of size $n$ from the GEV$(\mu, \sigma, \xi)$ distribution. The standardization is expressed as $$z=\dfrac{y- \Bar{y}}{\mathrm{IQR}},$$ where $\Bar{y}$ is the sample mean, $\mathrm{IQR}$ is the sample interquartile range, and $z$ is the standardized GEV sample. It is crucial to understand that standardizing a GEV sample does not guarantee that the standardized sample will follow a GEV distribution. However, rescaling the sample can alter its location ($\mu$) and scale ($\sigma$). The adjustment of $\mu$ is given by $$\dfrac{\mu-\Bar{y}}{IQR}$$ and the adjustment of $\sigma$ is $$\dfrac{\sigma}{IQR}.$$ To account for these changes, we train $\mathcal{N}$  with transformed percentile values and then invert the transformation to the original scale to compare with the true values.

By implementing standardization, a NN can become more robust and generalize better to extreme events, such as precipitation and wind, measured in different units, making it more versatile. Standardization is typically performed using pairing methods such as median with interquartile range (IQR) or mean with standard deviation. In this work, we have opted for the sample mean-IQR pairing, as it yields better outcomes than other combinations. 

\section{Simulation Study}
\label{sec: sim}
This section presents the results of simulation studies on the precision of the neural model $\mathcal{N}$ using a test set of $N_{test}$ parameter configurations generated from the selected parameter range described in Section~\ref{subsec:train}.

\subsection{Comparison with ML approach}
\label{subsec:CompareMLE}
For comparison of the model performance, we also calculate the MLEs of the GEV parameters on the test dataset. The MLEs are computed using the parameter estimation method implemented in the R package \texttt{ismev} (\cite{stephenson2011package}). In particular, the \texttt{ismev} package uses the \texttt{optim} function for numerical optimization to provide the MLEs of GEV distribution using the Nelder-Mead optimization method (see, \cite{singer2009nelder}). The accuracy of the parameter estimates from $\mathcal{N}$ is presented in Figure~\ref{fig:2}, Figure~\ref{fig:3}, along with the outcomes from the ML method.

\begin{figure}[p]
\centering
\hspace{-1cm}\includegraphics[width=0.95\textwidth]{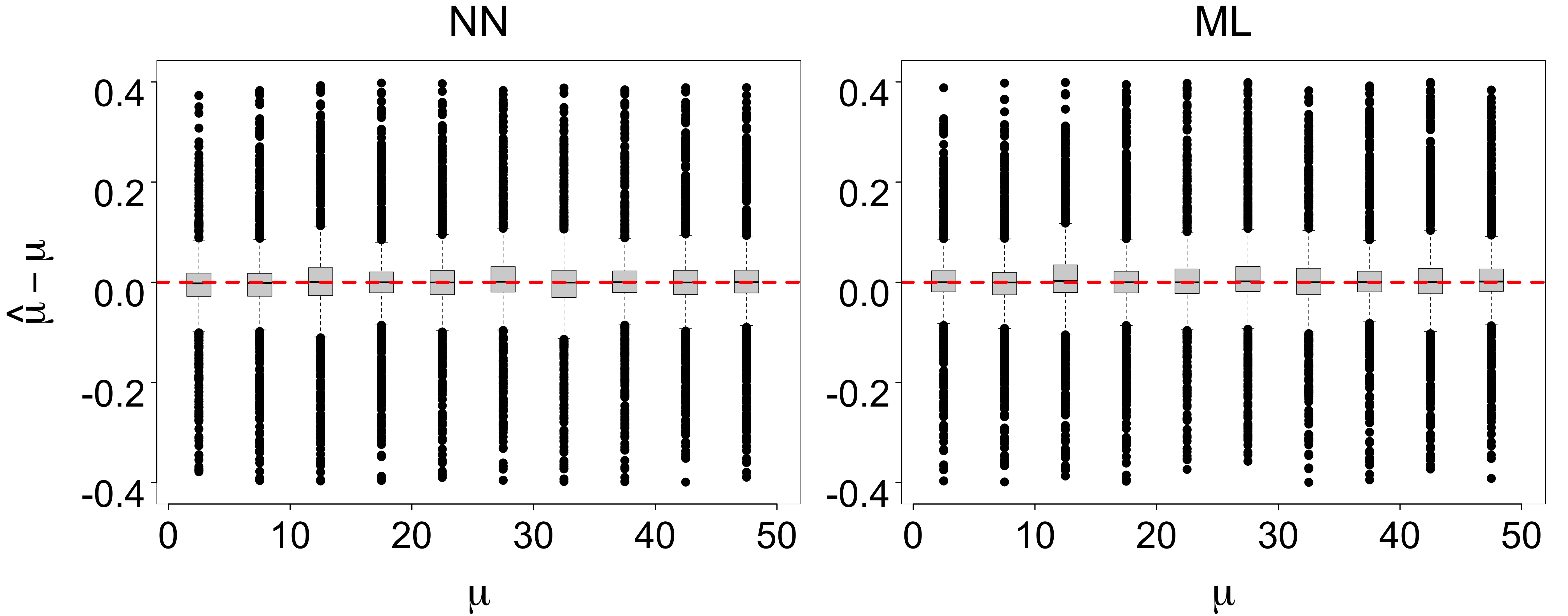}\\
\hspace{-1cm}\includegraphics[width=0.95\textwidth]{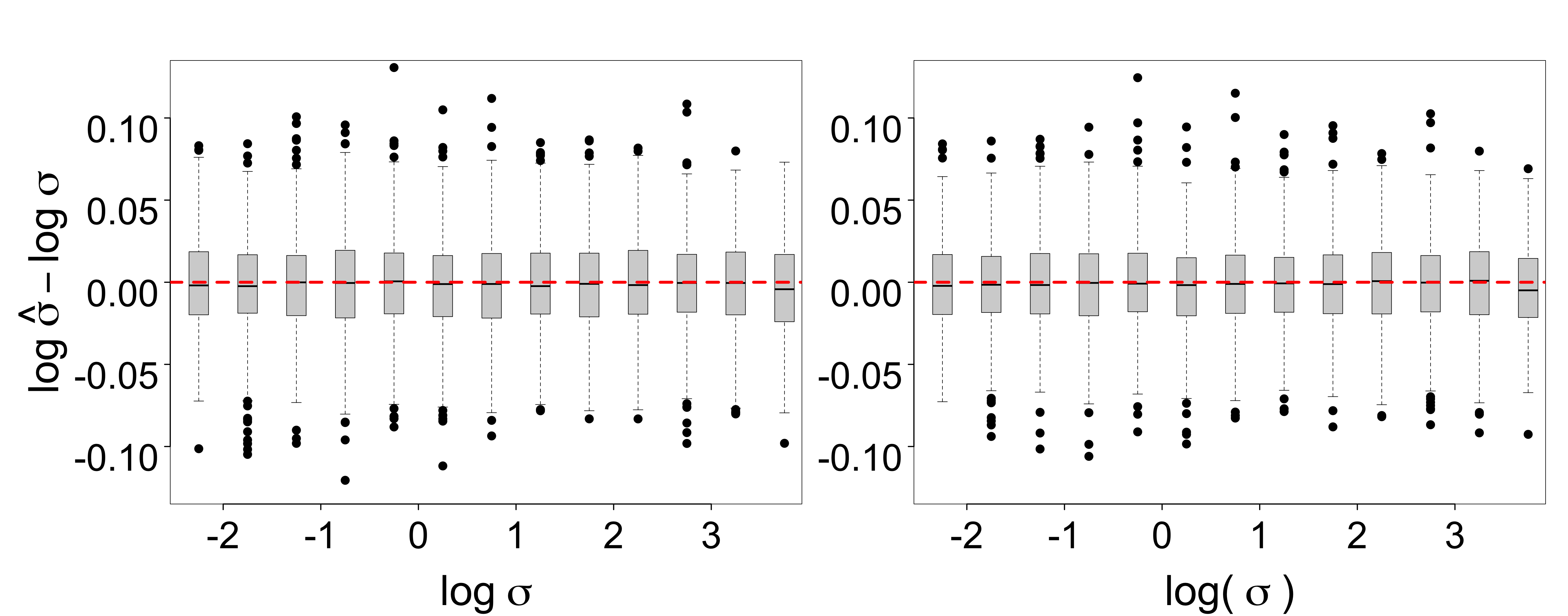}\\
\hspace{-1cm}\includegraphics[width=0.95\textwidth]{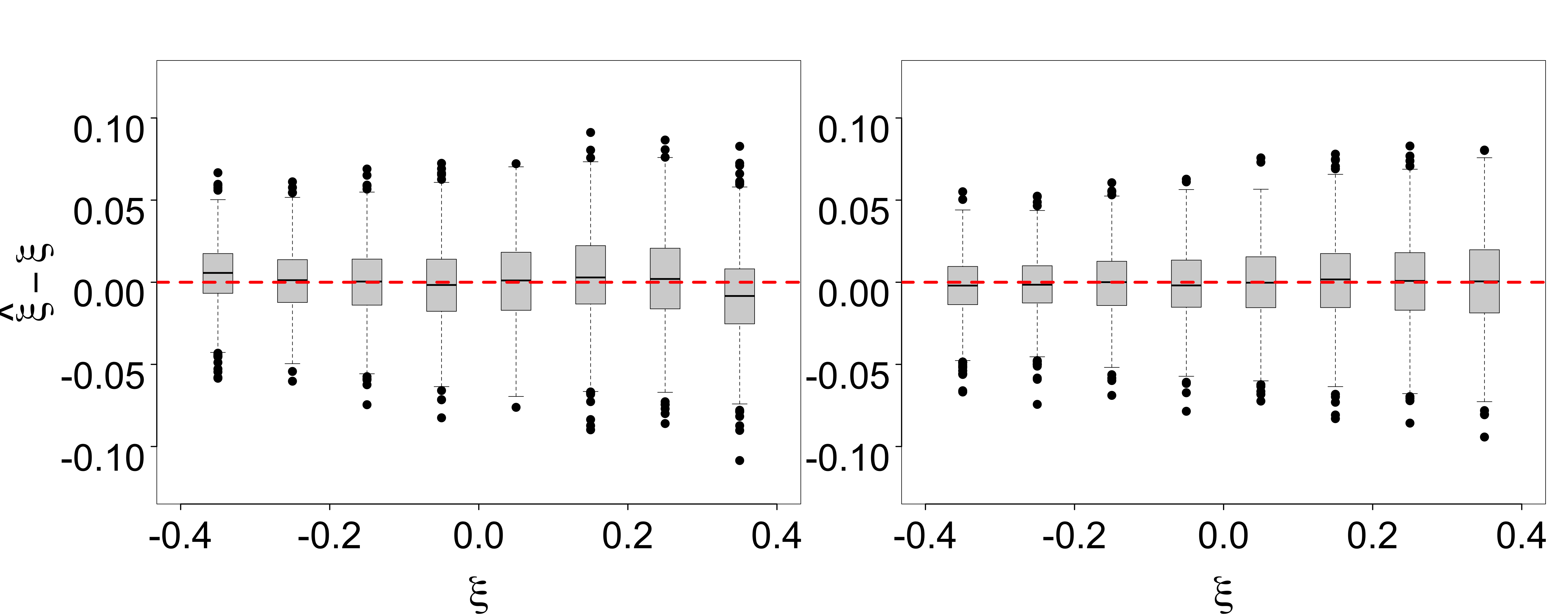}\\
\caption{This figure summarizes the results of our simulation study using a fixed GEV sample of size 1000. The boxplot depicts the deviation of the estimates $\widehat \mu$, log $ \widehat\sigma$, and $\widehat \xi$ from their true values over a 10,000 test set. The x-axis displays the parameter's true value, while the y-axis represents the deviation in the estimate from the true value.}
\label{fig:2}
\end{figure}

\begin{figure}[p]
\centering
\includegraphics[width=0.98\textwidth]{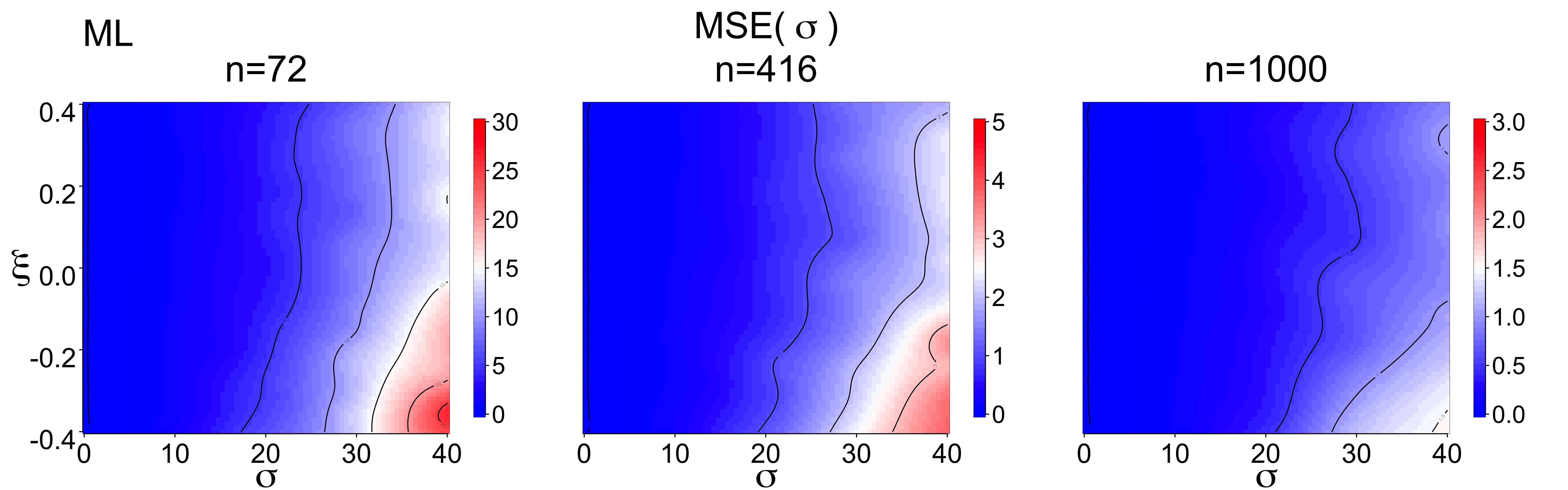}\\
\vspace{-0.05cm}
\includegraphics[width=0.98\textwidth]{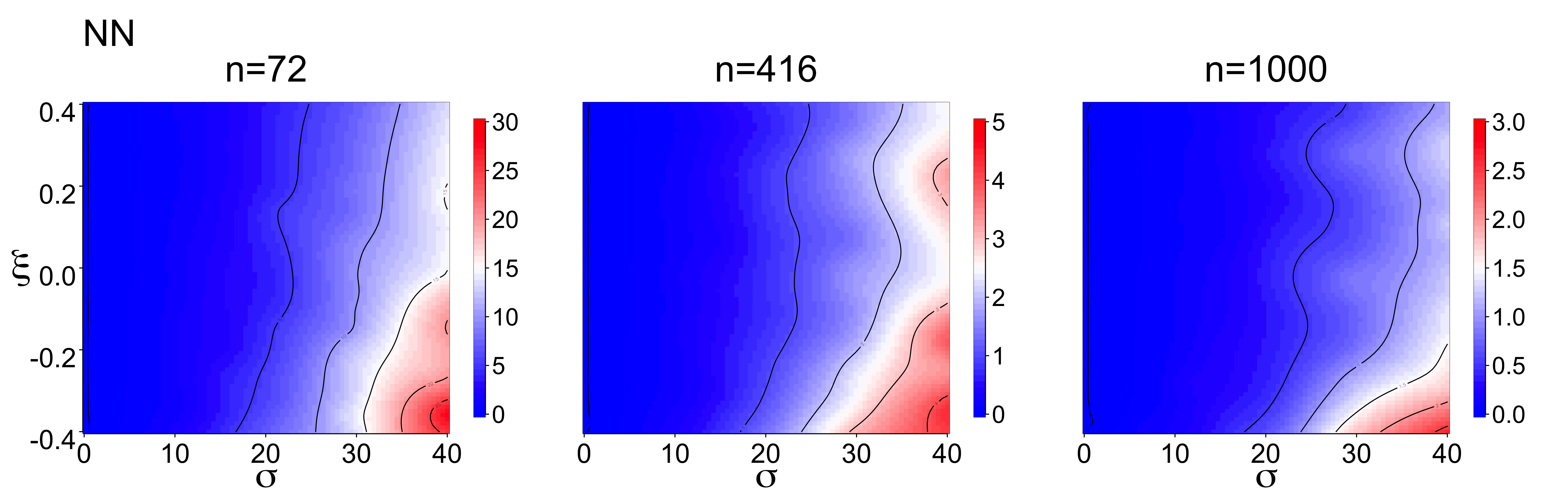}\\ 
\vspace{0.1cm}
\includegraphics[width=0.98\textwidth]{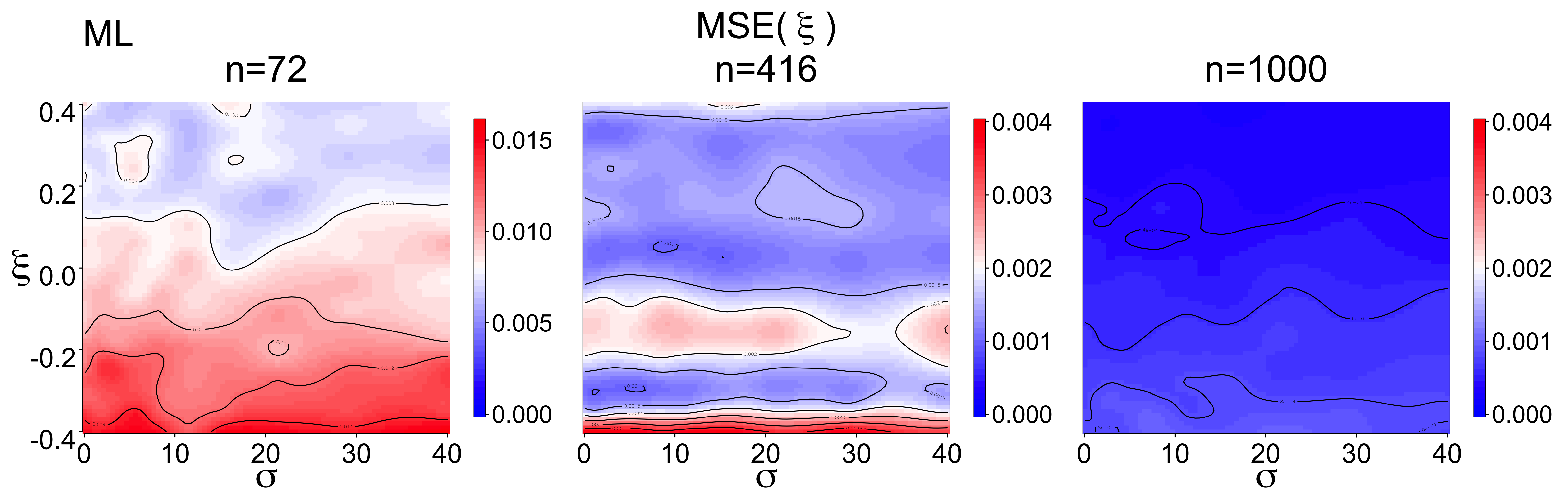}\\ 
\vspace{-0.05cm}
\includegraphics[width=0.98\textwidth]{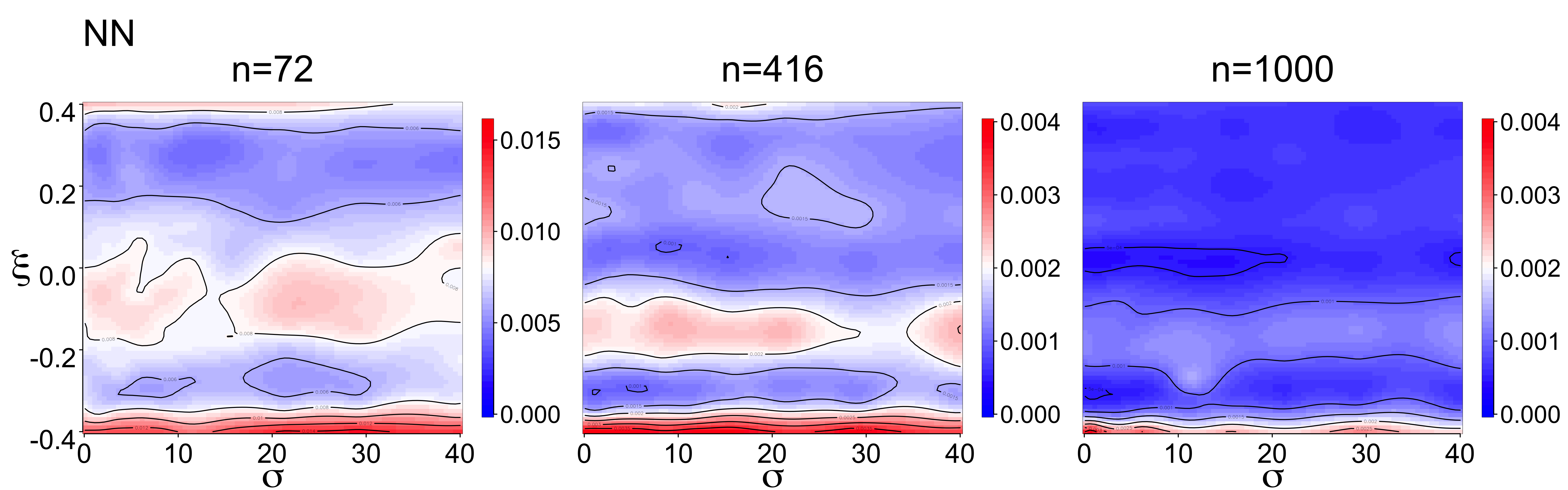}\\
\caption{This figure explains the behavior of the MSE as sample size varies among {72, 416, 1000}, based on our simulation study. The image plot showcases the MSE across the true $(\sigma, \xi)$, with $\sigma$ on the x-axis and $\xi$ on the y-axis.}
\label{fig:3}
\end{figure}

For our simulation study, case (1), we assess the performance of our neural model $\mathcal{N}$ by running $N_{test}=10,000$ configurations, each with 1000-GEV samples. To compare our model's estimates with MLEs, we use boxplots to visualize the differences between the estimates and the true parameter values across a range of true parameter values. Our analysis shows that the NN estimator performed similarly to the ML approach, with fewer outliers. These findings suggest that our neural model $\mathcal{N}$ is a promising approach for estimating GEV parameters.

In case 2, we use the same network architecture and parameter configuration for training and validation as in the previous case. To evaluate the behavior of the model, we generate test sets over a parameter grid of size $20 \times 20$ of $(\sigma, \xi)$ with $\mu=0$. For each configuration, we replicate the test sets 100 times; this is done across each sample size. In Figure~\ref{fig:3}, we compare the mean squared error (MSE) of the estimates obtained from the NN model and the ML method for the parameters $\sigma$ and $\xi$. We find that increasing the sample size from 72 to 1000 results in a decrease in MSE for both the NN and ML approaches. For a sample size of 72, the NN estimates have smaller MSE than the ML estimates, but for increasing sample sizes to 416 and 1000, the MLEs have smaller MSE than the NN estimates. However, the difference in MSE between the two approaches is small overall across the sample sizes.

Overall, this study provides evidence that NN models can serve as a viable alternative to the traditional ML approach for modeling extreme value. 

\subsection{Bootstrap}
\label{subsec:boot}
To account for uncertainty in parameter estimates obtained from the NN, we employ a parametric bootstrap approach (\cite{efron1994introduction}). We generate $B=900$ bootstrap samples from the original data, fit the NN to each sample, and compute a 95$\%$ CI of the true parameters. The bootstrap method incurs no additional computation costs after the NN is trained. We can produce B bootstrap replications and derive the corresponding results from the NN within seconds.

To evaluate the performance of the bootstrap-based CIs, we compare them to the likelihood-based CIs computed using the standard errors of the MLEs from the Hessian matrix of the maximum likelihood approach. We use the \texttt{ismev} package, as described in section \ref{subsec:CompareMLE}, for the computation. We compute the maximum likelihood-based CIs over the test set and obtain the bootstrap-based CIs using the fixed 1000 GEV sample-trained NN over 10,000 test sets and 900 bootstrap replications. 

Figure~\ref{fig:4} presents the ratio of the bootstrap-NN CI widths to the ML-based CI widths across the true parameter values. The boxplot summarizes the spread of this ratio, and our results indicate that the bootstrap-NN CIs are slightly wider than those of the ML model, although the observed difference is minimal. Specifically, we found that the bootstrap CIs for $\mu$ and $\xi$ are wider than the ML CIs, compared to $\sigma$. The CI width obtained from the bootstrap method may be wider than that obtained from the ML method because the bootstrap does not assume strong distributional constraints.

\begin{figure}[!ht]
\centering
\hspace{-0.8cm}\includegraphics[width=1.04\linewidth]{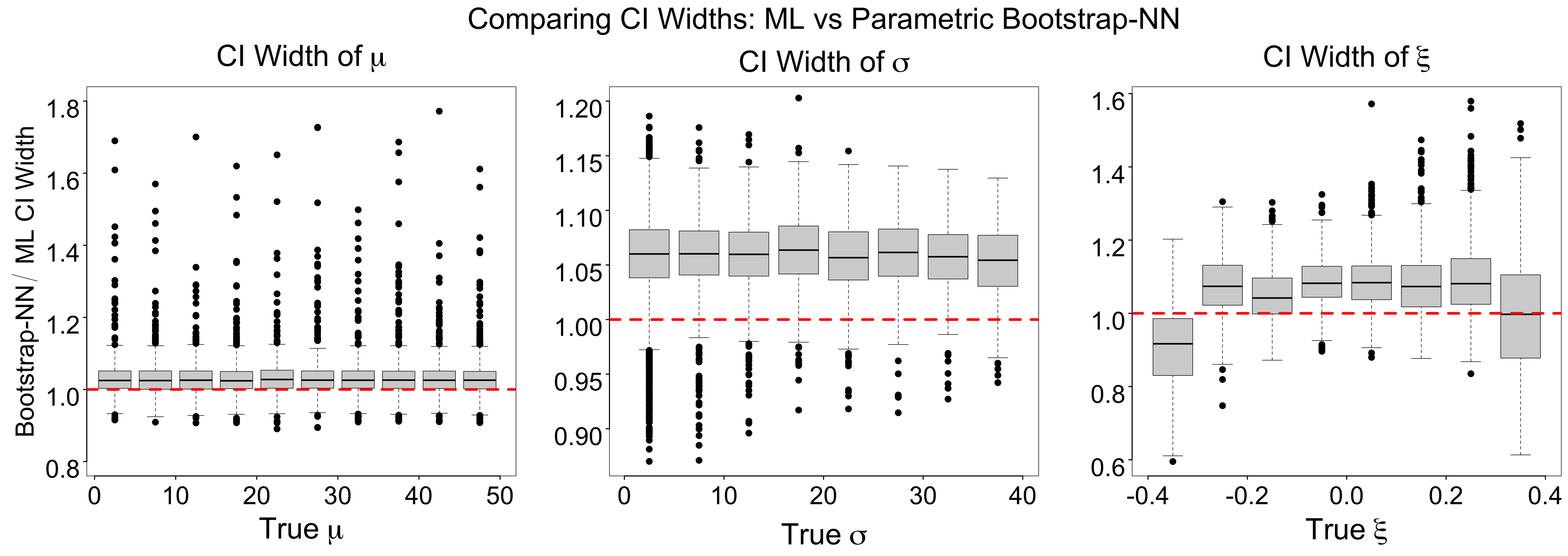}\\
\caption{The figure compares the CI widths obtained from the Bootstrap-NN and ML approaches for the GEV parameters across 10,000 test sets. We use the fixed 1000-sample trained NN  with 900 bootstrap replications to compute the bootstrap-NN output. The x-axis represents the true parameter values, and the y-axis shows the ratio of CI width from Bootstrap-NN to ML.}
\label{fig:4}
\end{figure}

\subsection{Timing comparison}
The training of the neural model $\mathcal{N}$ was found to be computationally efficient in comparison to other estimation methods. The model was implemented on the cloud-based Python platform, Google Colab, utilizing a computing environment with 2 virtual CPUs, 32GB of RAM, and either a P100 GPU with 16GB of memory or a T4 GPU with 16GB of GPU memory and system RAM that can be expanded up to 25GB. For comparison purposes, the MLEs were calculated using the R package ismev. The calculations were performed on a laptop with a 2.3 GHz Dual Core Intel i5 processor and 8GB of RAM. The evaluation time for the NN model on 10,000 test samples was 4 seconds, while the calculation of MLEs took approximately 10.631 minutes. Based on this, we anticipate a significant speed increase of over 150 times when scaling up to the target model output.

\section{Case Study}
\label{sec:casestudy}
Another way to validate and time out our NN estimator is to reproduce the results from a substantial climate model analysis. 
We analyze temperature extremes in the millennial runs of the Community Climate System Model version 3 (CCSM3), a global climate model widely used in climate research, at varying atmospheric $\mathrm{CO}_2$ concentrations (\cite{collins2006community}). The CCSM3 model includes complete representations of the atmosphere, land, sea ice, and ocean components, is run on grids of T31 resolution for the atmosphere and land, and approximately 1$^{\circ}$ resolution for the ocean and sea ice (Huang et al. 2016; Yeager et al. 2006).
\begin{figure}[!ht]
\centering
\includegraphics[width=\linewidth]{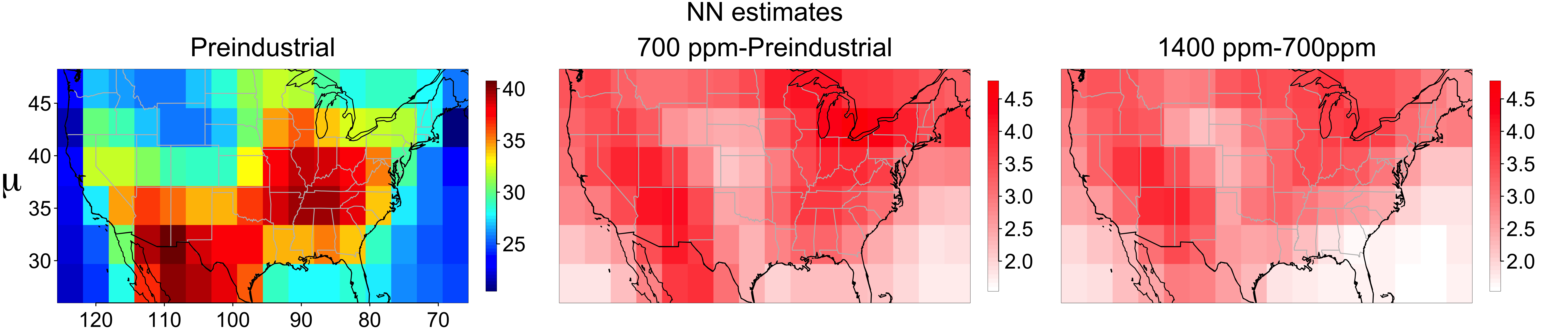}\\
\includegraphics[width=\linewidth]{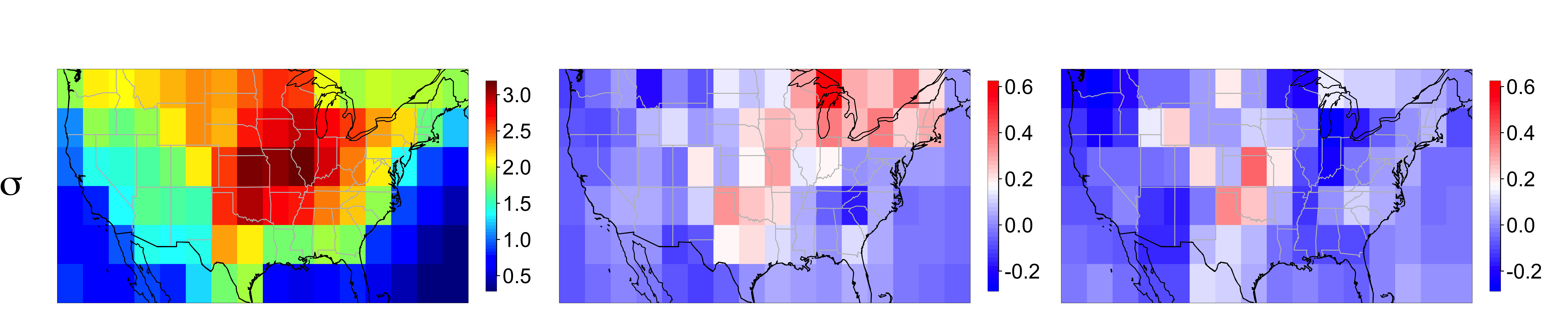}\\
\includegraphics[width=\linewidth]{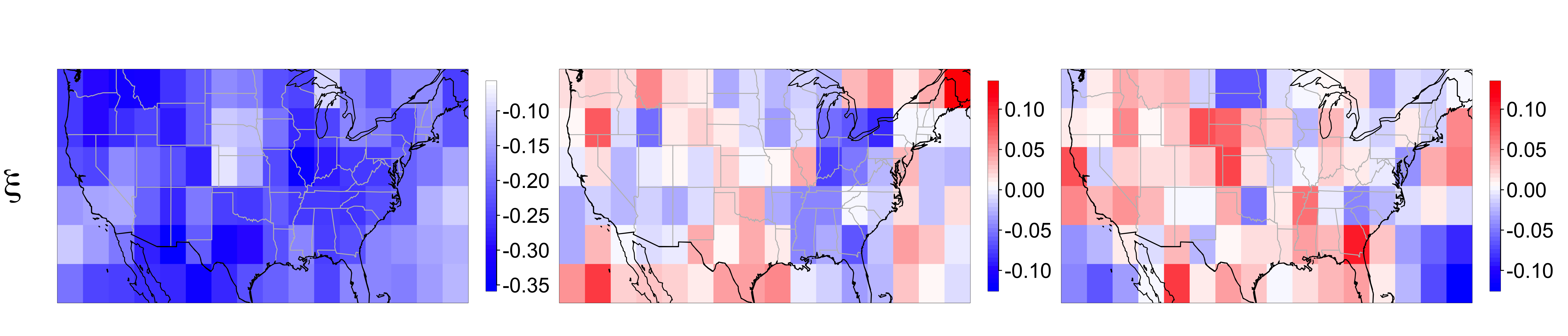}\\
\caption{NN estimates of CCSM3 GEV parameter for the pre-industrial period and possible changes for future cases. On the Left: are estimates for the pre-industrial center: the expected change in parameter estimates moving from 289 ppm $\mathrm{CO}_2$ to 700 ppm $\mathrm{CO}_2$ concentration, and right: is the change in estimates for 700 ppm $\mathrm{CO}_2$ to 1400 ppm.}
\label{fig:5}
\end{figure}

\begin{figure}[p]
\centering
\includegraphics[width=\linewidth]{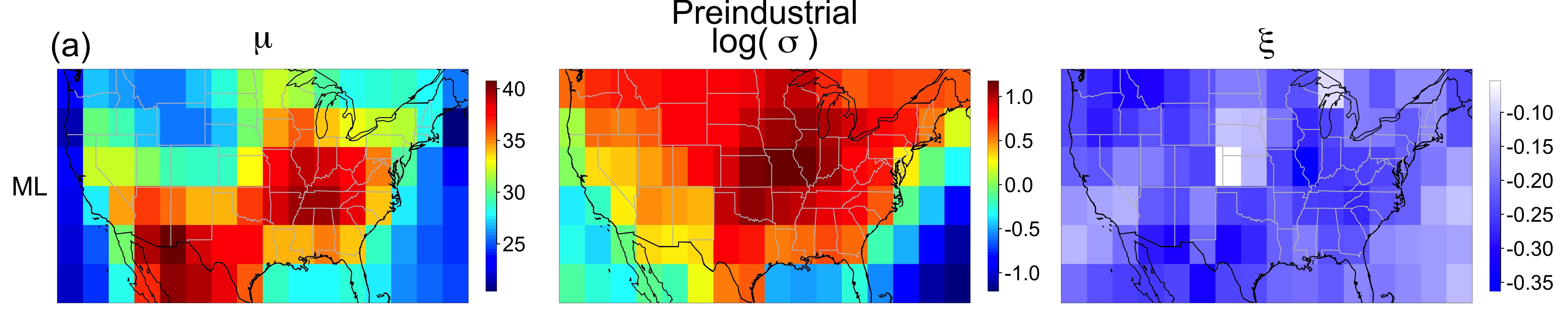}\\
\includegraphics[width=\linewidth]{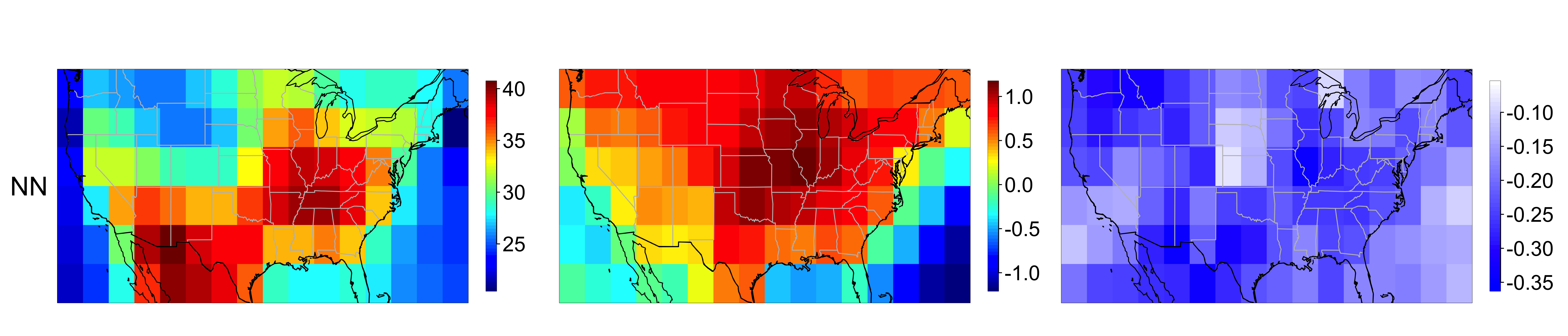}\\
\includegraphics[width=\linewidth]{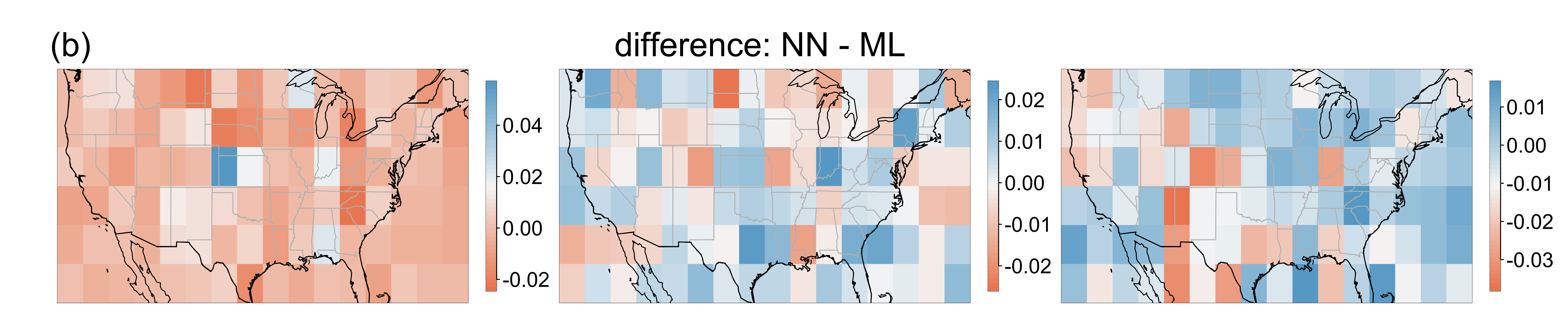}\\
\includegraphics[width=\linewidth]{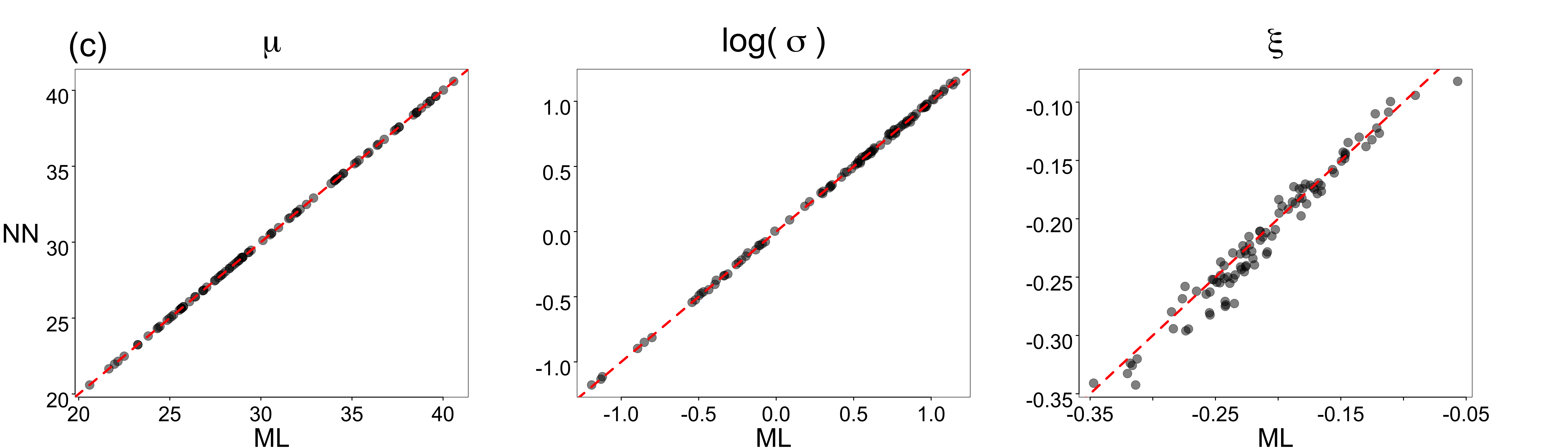}\\
\caption{(a) Show the CCSM3 GEV parameter estimates from the ML and the NN model across the spatial locations (b) Plot shows the difference in the parameter estimate values coming from ML and NN, and (c) Scatter plot to compare the outcomes of the NN vs ML over the GEV parameters.}
\label{fig:6}
\end{figure}

We consider a control run of 1000 years at 133 spatial locations across North America and consider three $\mathrm{CO}_2$ concentration scenarios: pre-industrial (289 ppm), future scenarios with 700 ppm (3.4$^{\circ}$C increase in global mean temperature) and 1400 ppm (6.1$^{\circ}$C increase in global mean temperature) (Huang et al. 2016). The key external forcings, including solar forcing and aerosol concentrations, are fixed at pre-industrial levels. The final simulations of 1000 years are assumed to be stationary and free from climate drift after a spin-up period. The maximum daily temperature is calculated for each grid box and year from the model output.

 We model the 1000-year annual maximum temperatures over the spatial domain using the GEV distribution. We fit a GEV distribution to each site in the domain, assuming each site has its specific GEV distribution over the 1000-year annual maxima values.
 To estimate the GEV parameters for each grid box, we use $\mathcal{N}$ as described in previous sections. Figure~\ref{fig:5} shows the NN estimates of CCSM3 GEV parameters for the pre-industrial period and possible changes for future cases. Our results show that the GEV distribution for the pre-industrial period is in agreement with previous findings from \cite{huang2016estimating}. Negative shape parameters are commonly observed when modeling extreme temperatures, and our model output confirms this trend. We further display the comparison of the performance of our NN model with the ML model in estimating the GEV parameters in Figure~\ref{fig:6}.
 
 Finally, using the same setup as in Section~\ref{subsec:boot}, we can compute a bootstrap-based confidence interval with 900 bootstrap replicates in approximately 0.4 seconds.

\section{Conclusion}
\label{sec:conc}
This study highlights advances in the use of deep learning algorithms for likelihood-free inference. The results indicate that a well-trained NN can estimate the parameters of complex heavy-tailed distributions, such as the GEV, with accuracy comparable to traditional ML approaches. Although there may be more variability than MLE in the estimation of the shape parameter compared to other parameters, this is expected due to the challenging nature of estimating the shape in heavy-tailed GEV. Additionally, our findings demonstrate a significant increase in computational speed, with a factor of 150 improvements in model evaluation times when compared to traditional ML approaches when working with large datasets. The use of NNs allows us full control over the testing and training samples and the ability to operate on a wide range of parameters. This allows us to customize the NN to meet the specific requirements of our problem and assess its reliability.

However, several limitations of the NN approach must be taken into account for its use. The selection of appropriate hyperparameters is a critical step in building a NN, as it can significantly impact the performance and accuracy of the model. Hyperparameters such as the number of hidden layers, number of neurons in the hidden layer, choice of activation function, learning rate, and batch size must be carefully chosen through a trial and error process. This process can prove to be time-consuming and challenging due to the vast and complex search space of hyperparameters. Additionally, the optimization of weights over each layer in the NN model can result in a large number of parameters, making the model intractable. Furthermore, the selection of the parametric range for the design of the training set is of utmost importance. The choice of informative statistics used as inputs to the network must also be carefully considered, as they must provide sufficient information about the data to allow for accurate estimates.

In conclusion, it is imperative to consider the network architecture, hyperparameter selection, and choice of statistics when utilizing the proposed model to ensure reliable results. Our estimation of the GEV parameters has not taken into account any potential spatial or temporal effects. This opens up the possibility for future studies to examine the integration of time-dependent structures into the GEV parameter estimation, leading to improved accuracy and robustness in extreme value predictions. Furthermore, the expanded usage of this approach in the spatial modeling of extremes can provide valuable insights into the distribution and behavior of extreme events in various geographical locations. Also, this could be beneficial in other statistical modeling approaches related to a heavy-tailed distribution.\\

\section*{Acknowledgements}
\label{sec: ackn}
We extend our sincere gratitude to Whitney Huang for generously sharing the 1000 years of output from three multimillennial runs of the CCSM3 model for our case study.
\bigskip
\bibliographystyle{agsm}
\bibliography{bibliography.bib}

\end{document}